\documentclass[preprint,12pt,sort&compress]{elsarticle}

\usepackage{textcomp,booktabs}
\usepackage[usenames,dvipsnames]{color}
\usepackage{colortbl}
\definecolor{mygray}{gray}{.9}
\definecolor{mypink}{rgb}{.99,.91,.95}
\definecolor{mycyan}{cmyk}{.3,0,0,0}
\usepackage{amssymb}
\usepackage{graphicx}
\usepackage{booktabs}
\usepackage{tabularx}
\usepackage{array}
\usepackage{lscape}
\usepackage{longtable}
\usepackage{multirow}
\usepackage{amsmath}
\usepackage{colortbl}
\usepackage[figuresright]{rotating}
\usepackage{epsfig}
\usepackage{epsf}
\usepackage{subfigure}
\usepackage{threeparttable}

\newcommand{\PreserveBackslash}[1]{\let\temp=\\#1\let\\=\temp}
\newcolumntype{C}[1]{>{\PreserveBackslash\centering}p{#1}}
\newcolumntype{R}[1]{>{\PreserveBackslash\raggedleft}p{#1}}
\newcolumntype{L}[1]{>{\PreserveBackslash\raggedright}p{#1}}

\journal{} 
\begin{document}

\begin{frontmatter}



\title{A quantum dynamic belief model to explain the interference effects of categorization on decision making}


\author[address1]{Zichang He}
\author[address1]{Wen Jiang\corref{label1}}
\address[address1]{School of Electronics and Information, Northwestern Polytechnical University, Xi'an, Shaanxi, 710072, China}
\cortext[label1]{Corresponding author at: School of Electronics and Information, Northwestern Polytechnical University, Xi'an, Shaanxi 710072, China. Tel: (86-29)88431267. E-mail address: jiangwen@nwpu.edu.cn, jiangwenpaper@hotmail.com}

\begin{abstract}
Categorization is necessary for many decision making tasks. However, the categorization process may interfere the decision making result and the law of total probability can be violated in some situations. To predict the interference effect of categorization, some model based on quantum probability has been proposed. In this paper, a new quantum dynamic belief(QDB) model is proposed. Considering the precise decision may not be made during the process, the concept of uncertainty is introduced in our model to simulate real human thinking process. Then the interference effect categorization can be predicted by handling the uncertain information. The proposed model is applied to a categorization decision-making experiment to explain the interference effect of categorization. Compared with other models, our model is relatively more succinct and the result shows the correctness and effectiveness of our model.
\end{abstract}

\begin{keyword}
Quantum dynamic model; Uncertain information; Interference effect of categorization; Law of total probability; categorization decision-making experiment
\end{keyword}

\end{frontmatter}
\section{Introduction}\label{Introduction}
The fields of cognition and decision making are well-developed. As an important and normal part of decision making process, categorization is widely involved in decision making tasks in reality. Without a precise categorization many decisions will not be able to make. For instance, doctors must classify the tumor as a benign one or a malignant one before doing the surgery; judges need to categorize the defendant as guilty or innocent before making a judgement; unexpected aircrafts need to be categorized as an enemy or not before commanders make decision. Categorization is regarded as a necessary part during the decision making, however, the effect of categorization is rarely studied separately.

Actually, a lot of living examples and experiments have shown that the categorization will bring about interference effects on decision making. The law of total probability will be broken due to the interference effect like categorization decision-making experiment\cite{townsend2000}. In order to explain the phenomenon and predict the interference effect, some models have been proposed like the quantum Belief-action entanglement(BAE) model proposed by Zheng Wang and Busemeyer\cite{wang2016interference}, an exemplar model account of feature inference proposed by Nosofsky R.M\cite{Nosofsky2015An} and so on. Quantum information has a wide application, like user security\cite{yu2015enhancing,Sharma2016A}, quantum communication\cite{song2015finite,Gisin2015Quantum} and so on. Quantum theory has been widely applied in the field of cognition and decision\cite{Khrennikov2004Information,aerts2015unreasonable,Yukalov2015Quantum,blutner2013quantum}. Quantum probability theory can solve many paradoxes like the violation of sure thing principle\cite{khrennikov2009quantum,cheon2010interference}, the additive law of probability\cite{brainerd2015episodic}, the Ellsberg paradox paradox\cite{Alnowaihi2016The} and many judgement errors\cite{busemeyer2011quantum}. The quantum model can also be adopted to explain many phenomena which are irregular in classical theories like order effect\cite{trueblood2011quantum,wang2013quantum,wang2014context}, disjunction effect\cite{Pothos2009A}, interference effect\cite{Nyman2011Quantum,busemeyer2007quantum} and so on. Among them, disjunction effect and interference effect are widely studied. Many models have been proposed, such as quantum dynamical model\cite{Pothos2009A,busemeyer2009empirical}, quantum-like approach\cite{Nyman2011On,Nyman2011Quantum}, quantum prospect decision theory\cite{Yukalov2015Quantum,Yukalov2011Decision} and quantum-like Bayesian networks\cite{Moreira2016Quantum} etc. Besides, quantum theory can explain some classical phenomena like prisoner's dilemma\cite{busemeyer2012social,chen2003quantum}. And it is also widely applied in game theory\cite{Makowski2009Transitivity,Alonso2015A,Alonso2016On}.

In this paper, a quantum dynamic belief(QDB) model which introduces an uncertain part to the decision making is proposed to predict the interference effects of categorization on decision making. Quantum dynamic model first proposed by Busemeyer $et.al$ in 2006\cite{busemeyer2006quantum} is formulated as random walk decision process. The evolution of complex valued probability amplitudes over time can be described in the dynamic model\cite{busemeyer2009empirical,Pothos2009A}. Decision making and optimization under uncertain environment is normal in reality and is heavily studied\cite{deng2015Generalized,Du2015832,2016AkyarFuzzy}. Uncertainty information modeling and processing is still an open issue\cite{You2012New,dengentropy,Vourdas2014Quantum}.
It is reasonable to assume that uncertainty exists in the decision making process. For example, in the test you are required to make a choice between plan A or plan B, you may be confused to make a precise choice, however the final decision is demanded. Considering it, the concept of uncertainty is introduced in our model to simulate the real human thinking process. A categorization decision-making experiment will be used to verify the correctness and effectiveness of the new model. The uncertain information will be handled differently in decision making along(D along) condition and categorization-decision making(C-D) condition, which will predict the interference effect well. Additionally, the new model is more succinct than classical quantum belief-action entanglement(BAE) model.

The rest of the paper is organized as follows. In Section 2, the preliminaries of the basic theory employed will be briefly introduced. And a categorization decision-making experiment which violates the law of total probability is illustrated in Section 3. Then our QDB model is proposed to predict the interference effects and explain the results of the experiment in Section 4. The model result and comparison is shown in Section 5. Finally, Section 6 comes to the conclusion.
\section{Preliminaries}
\subsection{quantum dynamic model}
The quantum dynamic model assumes that at the beginning of a game, a participant is possible to be in every state, thus the initial belief state is a superposition of all possible states in the form of a vector. The term $\psi $ represents the quantum probability amplitude of the according state. The term ${{{\left| {{\psi _{\rm{n}}}} \right|}^{\rm{2}}}}$ is the probability of observing state ${{\psi _{\rm{n}}}}$ initially.
\[\psi \left( {\rm{0}} \right){\rm{ = }}\frac{{\rm{1}}}{{\sqrt {{{\left| {{\psi _{\rm{1}}}} \right|}^{\rm{2}}}{\rm{ + }}{{\left| {{\psi _{\rm{2}}}} \right|}^{\rm{2}}}{\rm{ + }} \cdots {\rm{ + }}{{\left| {{\psi _{\rm{n}}}} \right|}^{\rm{2}}}} }}\left[ {{\psi _{\rm{1}}},{\psi _2}, \ldots ,{\psi _n}} \right]\]
During the decision making process, the belief state will evolve across time obeying a
Schr${\ddot o}$dinger equation.
\begin{equation}
\frac{d}{{dt}}\psi \left( t \right) =  - i \cdot H \cdot \psi \left( t \right)
\end{equation}
which has a matrix exponential solution
\begin{equation}\label{psit_2}
\psi \left( {{t_2}} \right) = {e^{ - iHt}} \cdot \psi \left( {{t_1}} \right)
\end{equation}
where $H$ is a Hamiltonian matrix $t$ to ensure a unitary matrix $U = {e^{ - iHt}}$ which ensures the sum of probabilities of all states always equals to 1.

For $t = {t_2} - {t_1}$, the transition probabilities ${T_{ij}}$ is the square of the unitary matrix ${U_{ij}}$.
\begin{equation}
{T_{ij}}\left( t \right) = {\left| {{U_{ij}}\left( t \right)} \right|^2}
\end{equation}
which means the probability of observing state $i$ at time ${t_2}$ given that state $j$ was observed at time ${t_1}$.

Based on the above definition, Eq.(\ref{psit_2}) could be rewritten as
\begin{equation}\label{psit_new}
\psi \left( t \right) = U \cdot \psi \left( 0 \right)
\end{equation}
The person's belief state evolves to $\psi \left( t \right)$ from the initial $\psi \left( 0 \right)$ across time $t$, which shows the dynamic process of the decision making. This model can produce interference effect which is impossible in classical Markov model\cite{busemeyer2009empirical}.
\subsection{Pignistic probability transformation}
Pignistic probability is widely applied to the field of decision making. The term "pignistic" proposed by Smets is originated from the word pignus, meaning 'bet' in Latin. Principle of insufficient reason is used to assign the basic probability of multiple-element set to singleton set. In other word, a belief interval is distributed into the crisp one determined as\cite{Smets1994The}:
\begin{equation}\label{PPT}{
bet\left( {{A_i}} \right) = \sum\nolimits_{{A_i} \subseteq {A_k}} {\frac{{m\left( {{A_k}} \right)}}{{\left| {{A_k}} \right|}}}
}\end{equation}
where ${\left| {{A_k}} \right|}$ denotes the number of elements in the set called the cardinality. Eq.(\ref{PPT}) is also called as Pignistic Probability Transformation(PPT).
\section{Categorization decision-making experiment}
\subsection{Experiment introduction}
Townsend $et.al$(2000)\cite{townsend2000} proposed a paradigm to study the interactions between categorization and decision making. The paradigm was designed to test Markov model initially. Subsequently, this paradigm was extended for quantum models as well by Busemeyer $et.al$(2009)\cite{busemeyer2009empirical}.
The paradigm consists of two kinds of conditions, one is categorization decision-making(C-D) condition and the other is decision along (D along) condition.
In C-D condition, in each trial, participants are shown pictures of faces, which vary along two dimensions(face width and lip thickness) like Figure \ref{face}.
\begin{figure}[!ht]
\centering
\includegraphics[scale=1]{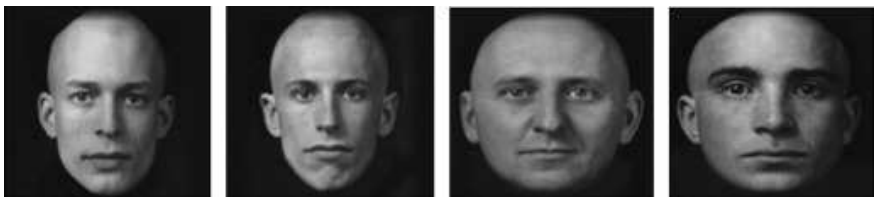}
\caption{Example faces used in a categorization-decision experiment}\label{face}
\end{figure}
Participants are asked to categorize the face as a "good" guy or a "bad" guy and then make a decision to "attack" or to "withdraw". The faces can be roughly divided into two kinds: one kind is "narrow" faces with narrow width and thick lips; the other kind is "wide" faces with wide width and thin lips. The participates are informed that "narrow" faces have a 0.60 probability of belonging to the "bad guy" population and "wide" faces had a 0.60 probability of belonging to the "good guy" population. And participants are rewarded for attacking "bad guy" and withdrawing from "good guy". In D along condition, the participants are asked to make a decision to attack or to withdraw directly without categorizing. The faces shown to them are the same as those in C-D condition. The whole experiment included a total of 26 participants, but each participant provides 51 observations for C-D condition for a total of ${\rm{26}} \times {\rm{51 = 1326}}$ observations, while each person produces 17 observations for D along condition for a total of ${\rm{17}} \times {\rm{26 = 442}}$ observations.
\subsection{Experiment results}
The experiment results are shown in Table \ref{classical result}.
\begin{table}[!h]
\centering
\caption{The results of C-D condition and D along condition}
\label{classical result}
\begin{tabular}{cccccccc}
\toprule
Face type & \textbf{$P\left( {G} \right)$} & \textbf{$P\left( {A|G} \right)$} & \textbf{$P\left( {B} \right)$} & \textbf{$P\left( {A|B} \right)$} & \textbf{${P_T}$} & \textbf{$P\left( {A} \right)$} &t \\
\midrule
Wide               & 0.84          & 0.35            & 0.16          & 0.52            & 0.37       & 0.39    &0.5733      \\
Narrow             & 0.17          & 0.41            & 0.83          & 0.63            & 0.59       & 0.69    &2.54        \\
\bottomrule
\end{tabular}
\end{table}
The column labeled $P\left( {G} \right)$ represents the probability of categorizing the face as a "good buy", the column labeled $P\left( {A|G} \right)$ represents the probability of attacking given categorizing the face as a "good guy". The column labeled $P\left( {B} \right)$ represents the probability of categorizing the face as a "bad buy". The column labeled $P\left( {A|B} \right)$ represents the probability of attacking given categorizing the face as a "bad guy". And the column labeled ${P_T}$ represents the final probability of attacking in C-D condition which is computed as follows.
\begin{equation}\label{Pt}
P\left( A \right) = P(G) \cdot P\left( {A|G} \right) + P\left( B \right) \cdot P\left( {A|B} \right)
\end{equation}
Accordingly, the column label as $P\left( {A} \right)$ represents the probability of attacking in D along condition. As shown in Table \ref{classical result}, there exist some deviation between ${P_T}$ and $P\left( {A} \right)$ for both types of face. However, the prominent deviation occurs as the narrow type faces were shown, which produces positive interference effect, while the interference effect is weak for the wide type faces. Using a paired $t$-test to test the significance of the difference between ${P_T}$ and $P\left( {A} \right)$ estimated from the 26 participants, the results shown in the $t$ column indicated that the interference effect is statistically significant for the narrow faces, but not for the wide faces.

The classical paradigm has been discussed in many works, the literatures of studying the categorization decision-making experiment and their results are shown below in Table \ref{preworks}.
\begin{table}[!h]
\centering
\small
\caption{Works of literature which study the categorization decision-making experiment}
\label{preworks}
\begin{threeparttable}
\begin{tabular*}{\columnwidth}{ccccccccc}
\toprule
Literature                                                                                           & Type & $P\left( {G} \right)$ & $P\left( {A|G} \right)$ & $P\left( {B} \right)$ & $P\left( {A|B} \right)$ & ${P_T}$   & $P\left( {A} \right)$   \\
\midrule
\multirow{2}{*}{\scriptsize\begin{tabular}[c]{@{}c@{}}Townsend $et.al$(2000)\cite{townsend2000}\end{tabular}}                     & W    & 0.84 & 0.35   & 0.16 & 0.52   & 0.37 & 0.39  \\
                                                                                                     & N    & 0.17 & 0.41   & 0.83 & 0.63   & 0.59 & 0.69   \\ \hline
\multirow{2}{*}{\scriptsize\begin{tabular}[c]{@{}c@{}}Busemeyer $et.al$(2009)\cite{busemeyer2009empirical}\end{tabular}}                    & W    & 0.80 & 0.37   & 0.20 & 0.53   & 0.40 & 0.39  \\
                                                                                                     & N    & 0.20 & 0.45   & 0.80 & 0.64   & 0.60 & 0.69   \\ \hline
\multirow{2}{*}{\scriptsize\begin{tabular}[c]{@{}c@{}}Wang and Busemeyer(2016)\\  Experiment 1\cite{wang2016interference}\end{tabular}} & W    & 0.78 & 0.39   & 0.22 & 0.52   & 0.42 & 0.42      \\
                                                                                                     & N    & 0.21 & 0.41   & 0.79 & 0.58   & 0.54 & 0.59   \\ \hline
\multirow{2}{*}{\scriptsize\begin{tabular}[c]{@{}c@{}}Wang and Busemeyer(2016) \\ Experiment 2\cite{wang2016interference}\end{tabular}} & W    & 0.78 & 0.33   & 0.22 & 0.53   & 0.37 & 0.37     \\
                                                                                                     & N    & 0.24 & 0.37   & 0.76 & 0.61   & 0.55 & 0.60   \\ \hline
\multirow{2}{*}{\scriptsize\begin{tabular}[c]{@{}c@{}}Wang and Busemeyer(2016) \\ Experiment 3\cite{wang2016interference}\end{tabular}} & W    & 0.77 & 0.34   & 0.23 & 0.58   & 0.40 & 0.39  \\
                                                                                                     & N    & 0.24 & 0.33   & 0.76 & 0.66   & 0.58 & 0.62   \\\hline
\multirow{2}{*}{Average}                                                                             & W    & 0.79 & 0.36   & 0.21 & 0.54   & 0.39 & 0.39      \\
                                                                                                     & N    & 0.21 & 0.39   & 0.79 & 0.62   & 0.57 & 0.64 \\
\bottomrule
\end{tabular*}
 \begin{tablenotes}
        \footnotesize
        \item[1] In Busemeyer $et.al$(2009)\cite{busemeyer2009empirical}, the classical experiment is replicated.
        \item[2] In Wang and Busemeyer(2016)\cite{wang2016interference}, Experiment 1 uses a larger data set to replicate the classical experiment. Experiment 2 introduce a new X-D trial verse C-D trial and this table only use the result of C-D trial. In experiment 3, the reward for attacking bad people is a bit less than the other two.
      \end{tablenotes}
\end{threeparttable}
\end{table}

According to the law of total probability, the probability of attacking should be the same in two conditions. Hence, the categorization brings about interference effect which makes a difference in decision making and breaks the law of total probability.
\section{Proposed method}\label{Proposed method}
\begin{figure}[!ht]
\centering
\includegraphics[scale=0.6]{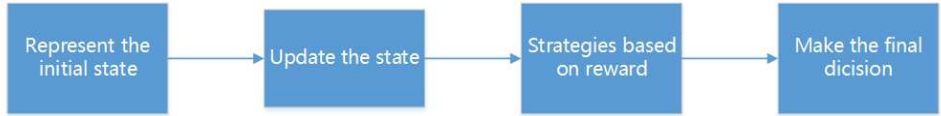}
 \caption{The model flow chart}\label{flowchart}
\end{figure}

In this section, our QDB model will be applied to explain and predict the experiment results shown in last section. To begin with, the integral flow chart of the model is shown in Figure \ref{flowchart}.

\textbf{Step 1. Represent the initial state}

In C-D condition, faces shown to participants are categorized as "good"(G) ones and "bad"(B) ones. Accordingly, the basic action options for participants is to withdraw(W) and to attack(A). However, in reality, it may be uncertain for some participants to make a precise choice. Considering it, a new temporary possible action labeled as uncertain(U) is brought into our model. Hence, the experiment involves a set of six mutual and exhaustive events
$\left\{ {AG,UG,WG,AB,UB,WB} \right\}$, where $AG$ symbolized the event that the participants decided to attack given the face was categorized as a good guy. Our model assumes that these six events correspond to six mutual and exhaustive states of the participant
$\left\{ {\left| {AG} \right\rangle ,\left| {UG} \right\rangle ,\left| {WG} \right\rangle ,\left| {AB} \right\rangle ,\left| {UB} \right\rangle ,\left| {WB} \right\rangle } \right\}$
It should be noticed that the states ${\left| {UG} \right\rangle }$ and ${\left| {UB} \right\rangle }$ are intermediate states whose probabilities will be divided into the other four states in the final decision making stage.

\begin{figure}[!ht]
\centering
\includegraphics[scale=0.5]{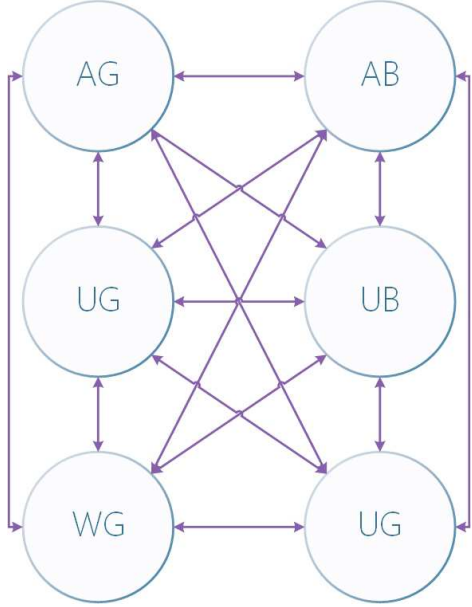}
 \caption{Transition among states}\label{circle}
\end{figure}

It is assumed that at the beginning of the experiment, a participant holds the potential to be in every possible state and the states can transit with each other like shown in Figure \ref{circle}, namely the initial state is totally random.
Hence, the participant's state is a superposition of the six basis states as follows.
\begin{equation}
\left| \psi  \right\rangle {\rm{ = }}{\psi _{AG}} \cdot \left| {AG} \right\rangle {\rm{ + }}{\psi _{UG}} \cdot \left| {UG} \right\rangle {\rm{ + }}{\psi _{WG}} \cdot \left| {WG} \right\rangle {\rm{ + }}{\psi _{AB}} \cdot \left| {AB} \right\rangle {\rm{ + }}{\psi _{UB}} \cdot \left| {UB} \right\rangle {\rm{ + }}{\psi _{WB}} \cdot \left| {WB} \right\rangle
\end{equation}
In the meantime, the initial state corresponds to an amplitude distribution represented by the ${\rm{6}} \times {\rm{1}}$ matrix.
\begin{equation}
\psi \left( 0 \right) = \left[ {\begin{array}{*{20}{c}}
{{\psi _{AG}}}\\
{{\psi _{UG}}}\\
{{\psi _{WG}}}\\
{{\psi _{AB}}}\\
{{\psi _{UB}}}\\
{{\psi _{WB}}}
\end{array}} \right]
\end{equation}
where, ${\left| {{\psi _{AG}}} \right|^{\rm{2}}}$ is the probability of observing state ${\left| {AG} \right\rangle }$ initially and the same as the others. It should be noticed that the sum of all states' probability must equal to one.
\begin{equation}\label{a'*a=1}
{\left| \psi  \right|^{\rm{2}}}{\rm{ = }}\psi ' \cdot \psi {\rm{ = 1}}
\end{equation}

\textbf{Step 2. Update the state}

As the experiment goes, more information is known to the participants. In C-D condition, the categorization of the faces is informed to the participants. Assume that new information at time ${{\rm{t}}_1}$ changes the initial state $\psi \left( 0 \right)$ which is at time $t$=0 into a new state $\psi \left( {{{\rm{t}}_1}} \right)$.

In C-D condition, if the face is categorized as "good", the amplitude distribution across states will change to
\begin{equation}
\psi \left( {{t_1}} \right) = \frac{1}{{\sqrt {{{\left| {{\psi _{AG}}} \right|}^2} + {{\left| {{\psi _{UG}}} \right|}^2} + {{\left| {{\psi _{WG}}} \right|}^2}} }}\left[ {\begin{array}{*{20}{c}}
{{\psi _{AG}}}\\
{{\psi _{UG}}}\\
{{\psi _{WG}}}\\
0\\
0\\
0
\end{array}} \right] = \left[ {\begin{array}{*{20}{c}}
{{\psi _G}}\\
0
\end{array}} \right]
\end{equation}
It means that the probabilities of being the states ${{\psi _{AB}}}$, ${{\psi _{UB}}}$ and ${{\psi _{WB}}}$ change into zero and the probabilities of being the sates ${{\psi _{AG}}}$, ${{\psi _{UG}}}$ and ${{\psi _{WG}}}$ are enlarged with the same proportion $\frac{1}{{\sqrt {{{\left| {{\psi _{AG}}} \right|}^2} + {{\left| {{\psi _{UG}}} \right|}^2} + {{\left| {{\psi _{WG}}} \right|}^2}} }}$ to meet Eq. \ref{a'*a=1}, where
${\sqrt {{{\left| {{\psi _{AG}}} \right|}^2} + {{\left| {{\psi _{UG}}} \right|}^2} + {{\left| {{\psi _{WG}}} \right|}^2}} }$ is the initial probability of categorizing the face as "good". And the $2 \times 1$ matrix
$\left[ {\begin{array}{*{20}{c}}
{{\psi _G}}\\
0
\end{array}} \right]$ is the brief form of the state given categorizing the face as "good".

If the face is categorized as "bad", the amplitude distribution across states will change to
\begin{equation}
\psi \left( {{t_1}} \right) = \frac{1}{{\sqrt {{{\left| {{\psi _{AB}}} \right|}^2} + {{\left| {{\psi _{UB}}} \right|}^2} + {{\left| {{\psi _{WB}}} \right|}^2}} }}\left[ {\begin{array}{*{20}{c}}
0\\
0\\
0\\
{{\psi _{AB}}}\\
{{\psi _{UB}}}\\
{{\psi _{WB}}}
\end{array}} \right] = \left[ {\begin{array}{*{20}{c}}
{{\psi _G}}\\
0
\end{array}} \right]
\end{equation}
It means that the probabilities of being the states ${{\psi _{AG}}}$, ${{\psi _{UG}}}$ and ${{\psi _{WG}}}$ change into zero and the probabilities of being the sates ${{\psi _{AB}}}$, ${{\psi _{UB}}}$ and ${{\psi _{WB}}}$ are enlarged with the same proportion $\frac{1}{{\sqrt {{{\left| {{\psi _{AB}}} \right|}^2} + {{\left| {{\psi _{UB}}} \right|}^2} + {{\left| {{\psi _{WB}}} \right|}^2}} }}$ to meet Eq.\ref{a'*a=1}, where
${\sqrt {{{\left| {{\psi _{AB}}} \right|}^2} + {{\left| {{\psi _{UB}}} \right|}^2} + {{\left| {{\psi _{WB}}} \right|}^2}} }$ is the initial probability of categorizing the face as "bad". And the $2 \times 1$ matrix
$\left[ {\begin{array}{*{20}{c}}
0\\
{{\psi _B}}
\end{array}} \right]$ is the brief form of the state given categorizing the face as "bad".

In D along condition, there is no new information to bring the change. So the amplitude distribution across states will keep the same as the initial one.
\begin{equation}
\footnotesize
\begin{array}{l}
\psi \left( {{t_1}} \right) = \psi \left( 0 \right) = \left[ {\begin{array}{*{20}{c}}
{\sqrt {{{\left| {{\psi _{AG}}} \right|}^2} + {{\left| {{\psi _{UG}}} \right|}^2} + {{\left| {{\psi _{WG}}} \right|}^2}}  \cdot {\psi _G}}\\
{\sqrt {{{\left| {{\psi _{AB}}} \right|}^2} + {{\left| {{\psi _{UB}}} \right|}^2} + {{\left| {{\psi _{WB}}} \right|}^2}}  \cdot {\psi _B}}
\end{array}} \right]\\
{\kern 1pt} {\kern 1pt} {\kern 1pt} {\kern 1pt} {\kern 1pt} {\kern 1pt} {\kern 1pt} {\kern 1pt} {\kern 1pt} {\kern 1pt} {\kern 1pt} {\kern 1pt} {\kern 1pt} {\kern 1pt} {\kern 1pt} {\kern 1pt} {\kern 1pt} {\kern 1pt} {\kern 1pt} {\kern 1pt} {\kern 1pt} {\kern 1pt} {\kern 1pt} {\kern 1pt} {\kern 1pt}  = \sqrt {{{\left| {{\psi _{AG}}} \right|}^2} + {{\left| {{\psi _{UG}}} \right|}^2} + {{\left| {{\psi _{WG}}} \right|}^2}} \left[ {\begin{array}{*{20}{c}}
{{\psi _G}}\\
0
\end{array}} \right] + \sqrt {{{\left| {{\psi _{AB}}} \right|}^2} + {{\left| {{\psi _{UB}}} \right|}^2} + {{\left| {{\psi _{WB}}} \right|}^2}} \left[ {\begin{array}{*{20}{c}}
0\\
{{\psi _B}}
\end{array}} \right]
\end{array}
\end{equation}
The equation shows that the initial state for the participants under unknown condition is formed by a weighted sum of the amplitude distributions for the two known conditions.

\textbf{Step 3. Strategies based on reward}

The actions taken are always based on the according reward. Hence, in the model, the decision maker must evaluate the rewards in order to select an appropriate action, which changes the previous state at time ${{t_1}}$ into a new state ${\psi \left( {{t_2}} \right)}$ at time ${{t_2}}$. The evolution of the state during the time period corresponds to the thinking process which leads to a decision.

Based on the quantum dynamic model, the state evolution obeys a Schr${\ddot o}$dinger equation driven by a $6 \times 6$ Hamiltonian matrix $H$:
\begin{equation}
\frac{d}{{dt}}\psi \left( t \right) =  - i \cdot H \cdot \psi \left( t \right)
\end{equation}
which has a matrix exponential solution for $t = {t_1} - {t_2}$.
\begin{equation}
\psi \left( {{t_2}} \right) = {e^{ - i \cdot H \cdot t}} \cdot \psi (t)
\end{equation}
The unitary matrix is defined by
\begin{equation}
U\left( t \right) = {e^{ - i \cdot H \cdot t}}
\end{equation}
which determines the transition probabilities according to ${T_{ij}}\left( t \right) = {\left| {{U_{ij}}\left( t \right)} \right|^2}$. Let the component of ${T_{ij}}$ denote as ${t_{ij}}$, which represents the probability of observing state $i$ at time ${t_2}$ given that state $j$ was observed at time ${t_1}$. To create a choice probability that reaches its maximum, the process time parameter $t$ is set as ${\pi  \mathord{\left/
 {\vphantom {\pi  2}} \right.
 \kern-\nulldelimiterspace} 2}$ in the model.

The Hamiltonian matrix $H$ is a Hermitian matrix, which satisfies ${H^ * } = H$, so that $U$ is a unitary matrix, which satisfies
\begin{equation}
{U^ * } \cdot U = I
\end{equation}
The property is the key to guarantee that the state $\psi \left( t \right)$ keeps a unit length. Assume
\begin{equation}
H = \left[ {\begin{array}{*{20}{c}}
{{H_G}}&0\\
0&{{H_B}}
\end{array}} \right]
\end{equation}
with
\begin{equation}
{H_G} = \left( {\begin{array}{*{20}{c}}
{{h_G}}&0&1\\
0&1&0\\
1&0&{ - {h_G}}
\end{array}} \right)
\end{equation}
and
\begin{equation}
{H_B} = \left( {\begin{array}{*{20}{c}}
{{h_B}}&0&1\\
0&1&0\\
1&0&{ - {h_B}}
\end{array}} \right)
\end{equation}
The $3 \times 3$ Hamiltonian matrix ${H_G}$ applies when the participant categorizes the face as "good", and the other $3 \times 3$ Hamiltonian matrix ${H_B}$ applies when the participant categorizes the face as "bad". The parameter ${h_G}$ is a function of the difference between the reward for attacking relative to withdrawing given categorizing the face as "good" and the parameter ${h_B}$ is a function of the difference between the reward for attacking relative to withdrawing given categorizing the face as "bad". According to the payoffs, the Hamiltonian matrix transforms the state probabilities to attacking, uncertainty or withdrawing.

Based on the above, we can obtain the participant's state at time ${t_2}$ as follows:
In C-D condition, if the face is categorized as "good", the state will be
\begin{equation}
\psi \left( {{t_2}} \right) = {e^{ - i \cdot H \cdot t}} \cdot \psi ({t_{\rm{1}}}){\rm{ = }}\left[ {\begin{array}{*{20}{c}}
{{e^{ - i \cdot {H_G} \cdot t}}}&0\\
0&{{e^{ - i \cdot {H_B} \cdot t}}}
\end{array}} \right] \cdot \left[ {\begin{array}{*{20}{c}}
{{\psi _G}}\\
0
\end{array}} \right] = {e^{ - i \cdot {H_G} \cdot t}} \cdot {\psi _G}
\end{equation}
If the face is categorized as "bad", the state will be
\begin{equation}
\psi \left( {{t_2}} \right) = {e^{ - i \cdot H \cdot t}} \cdot \psi ({t_{\rm{1}}}){\rm{ = }}\left[ {\begin{array}{*{20}{c}}
{{e^{ - i \cdot {H_G} \cdot t}}}&0\\
0&{{e^{ - i \cdot {H_B} \cdot t}}}
\end{array}} \right] \cdot \left[ {\begin{array}{*{20}{c}}
0\\
{{\psi _B}}
\end{array}} \right] = {e^{ - i \cdot {H_B} \cdot t}} \cdot {\psi _B}
\end{equation}
In D along condition, the state does not change at time ${t_1}$ and it will change into
\begin{equation}
\footnotesize
\begin{array}{*{20}{l}}
{\psi \left( {{t_2}} \right) = {e^{ - i \cdot H \cdot t}} \cdot \psi (0){\rm{ = }}\left[ {\begin{array}{*{20}{c}}
{{e^{ - i \cdot {H_G} \cdot t}}}&0\\
0&{{e^{ - i \cdot {H_B} \cdot t}}}
\end{array}} \right] \cdot \left[ {\begin{array}{*{20}{c}}
{\sqrt {{{\left| {{\psi _{AG}}} \right|}^2} + {{\left| {{\psi _{UG}}} \right|}^2} + {{\left| {{\psi _{WG}}} \right|}^2}}  \cdot {\psi _G}}\\
{\sqrt {{{\left| {{\psi _{AB}}} \right|}^2} + {{\left| {{\psi _{UB}}} \right|}^2} + {{\left| {{\psi _{WB}}} \right|}^2}}  \cdot {\psi _B}}
\end{array}} \right]}\\
\begin{array}{l}
{\kern 1pt} {\kern 1pt} {\kern 1pt} {\kern 1pt} {\kern 1pt} {\kern 1pt} {\kern 1pt} {\kern 1pt} {\kern 1pt} {\kern 1pt} {\kern 1pt} {\kern 1pt} {\kern 1pt} {\kern 1pt} {\kern 1pt} {\kern 1pt} {\kern 1pt} {\kern 1pt} {\kern 1pt} {\kern 1pt} {\kern 1pt} {\kern 1pt} {\kern 1pt} {\kern 1pt} {\kern 1pt} {\kern 1pt} {\kern 1pt} {\kern 1pt} {\kern 1pt} {\kern 1pt}  = \sqrt {{{\left| {{\psi _{AG}}} \right|}^2} + {{\left| {{\psi _{UG}}} \right|}^2} + {{\left| {{\psi _{WG}}} \right|}^2} \cdot } {e^{ - i \cdot {H_G} \cdot t}} \cdot {\psi _G}{\rm{ + }}\\
{\kern 1pt} {\kern 1pt} {\kern 1pt} {\kern 1pt} {\kern 1pt} {\kern 1pt} {\kern 1pt} {\kern 1pt} {\kern 1pt} {\kern 1pt} {\kern 1pt} {\kern 1pt} {\kern 1pt} {\kern 1pt} {\kern 1pt} {\kern 1pt} {\kern 1pt} {\kern 1pt} {\kern 1pt} {\kern 1pt} {\kern 1pt} {\kern 1pt} {\kern 1pt} {\kern 1pt} {\kern 1pt} {\kern 1pt} {\kern 1pt} {\kern 1pt} {\kern 1pt} {\kern 1pt} {\kern 1pt} {\kern 1pt} {\kern 1pt} {\kern 1pt} {\kern 1pt} {\kern 1pt} {\kern 1pt} {\kern 1pt} {\kern 1pt} {\kern 1pt} {\kern 1pt} \sqrt {{{\left| {{\psi _{AB}}} \right|}^2} + {{\left| {{\psi _{UB}}} \right|}^2} + {{\left| {{\psi _{WB}}} \right|}^2}}  \cdot {e^{ - i \cdot {H_B} \cdot t}} \cdot {\psi _B}
\end{array}
\end{array}
\end{equation}
at time ${t_2}$. As we can see, the ${t_2}$ state for the participants under unknown condition is still formed by a weighted sum of the amplitude distributions for the two known conditions.

\textbf{Step 4. Make the final decision}

After the previous three steps, the participants's state has been obtained. The last step is to calculate the probabilities according to the state. To address it, a measure matrix $M$ is introduced to pick out the needed state.
\begin{equation}\label{M=MG MB}
M = \left( {\begin{array}{*{20}{c}}
{{M_G}}&0\\
0&{{M_B}}
\end{array}} \right)
\end{equation}
where ${M_G}$ is the matrix used to pick out state given categorizing the face as "good" and ${M_B}$ is the matrix used to pick out state given categorizing the face as "bad".

As the uncertain state is an assumed intermediate state, it should be divided into attacking state and withdrawing state as follows based on Eq. (\ref{PPT}):
\begin{equation}
P\left( A \right) = P\left( W \right) = 0.5 \cdot P\left( N \right)
\end{equation}

In C-D condition, to pick out the state of attacking, we set
\begin{equation}
{M_A} = \left( {\begin{array}{*{20}{c}}
{{M_{GA}}}&0\\
0&{{M_{BA}}}
\end{array}} \right)
\end{equation}
with
\begin{equation}
{M_{GA}} = {M_{BA}} = \left[ {\begin{array}{*{20}{c}}
1&0&0\\
0&0&0\\
0&0&0
\end{array}} \right]
\end{equation}
and then the probability of attacking given categorizing the face as "good" equals to
\begin{equation}
\footnotesize
\begin{array}{l}
P\left( {A|G} \right){\rm{ = }}{\left\| {{M_A} \cdot {e^{ - i \cdot t \cdot H}} \cdot \psi \left( {{t_1}} \right)} \right\|^{\rm{2}}}{\rm{ = }}{\left\| {\left[ {\begin{array}{*{20}{c}}
{{M_{GA}}}&{\rm{0}}\\
{\rm{0}}&{{M_{BA}}}
\end{array}} \right] \cdot \left[ {\begin{array}{*{20}{c}}
{{e^{ - i \cdot t \cdot {H_G}}}}&{\rm{0}}\\
{\rm{0}}&{{e^{ - i \cdot t \cdot {H_B}}}}
\end{array}} \right] \cdot \left[ {\begin{array}{*{20}{c}}
{{\psi _G}}\\
0
\end{array}} \right]} \right\|^{\rm{2}}}\\
{\kern 1pt} {\kern 1pt} {\kern 1pt} {\kern 1pt} {\kern 1pt} {\kern 1pt} {\kern 1pt} {\kern 1pt} {\kern 1pt} {\kern 1pt} {\kern 1pt} {\kern 1pt} {\kern 1pt} {\kern 1pt} {\kern 1pt} {\kern 1pt} {\kern 1pt} {\kern 1pt} {\kern 1pt} {\kern 1pt} {\kern 1pt} {\kern 1pt} {\kern 1pt} {\kern 1pt} {\kern 1pt} {\kern 1pt} {\kern 1pt} {\kern 1pt} {\kern 1pt} {\kern 1pt} {\kern 1pt} {\kern 1pt} {\kern 1pt} {\kern 1pt} {\kern 1pt} {\kern 1pt} {\kern 1pt} {\kern 1pt} {\kern 1pt} {\kern 1pt} {\kern 1pt} {\kern 1pt} {\kern 1pt} {\kern 1pt} {\rm{ = }}{\left\| {{M_{GA}} \cdot {e^{ - i \cdot t \cdot {H_G}}} \cdot {\psi _G}} \right\|^{\rm{2}}}
\end{array}
\end{equation}
The probability of attacking given categorizing the face as "bad" equals to
\begin{equation}
\footnotesize
\begin{array}{l}
P\left( {A|B} \right){\rm{ = }}{\left\| {{M_A} \cdot {e^{ - i \cdot t \cdot H}} \cdot \psi \left( {{t_1}} \right)} \right\|^{\rm{2}}}{\rm{ = }}{\left\| {\left[ {\begin{array}{*{20}{c}}
{{M_{GA}}}&{\rm{0}}\\
{\rm{0}}&{{M_{BA}}}
\end{array}} \right] \cdot \left[ {\begin{array}{*{20}{c}}
{{e^{ - i \cdot t \cdot {H_G}}}}&{\rm{0}}\\
{\rm{0}}&{{e^{ - i \cdot t \cdot {H_B}}}}
\end{array}} \right] \cdot \left[ {\begin{array}{*{20}{c}}
0\\
{{\psi _B}}
\end{array}} \right]} \right\|^{\rm{2}}}\\
{\kern 1pt} {\kern 1pt} {\kern 1pt} {\kern 1pt} {\kern 1pt} {\kern 1pt} {\kern 1pt} {\kern 1pt} {\kern 1pt} {\kern 1pt} {\kern 1pt} {\kern 1pt} {\kern 1pt} {\kern 1pt} {\kern 1pt} {\kern 1pt} {\kern 1pt} {\kern 1pt} {\kern 1pt} {\kern 1pt} {\kern 1pt} {\kern 1pt} {\kern 1pt} {\kern 1pt} {\kern 1pt} {\kern 1pt} {\kern 1pt} {\kern 1pt} {\kern 1pt} {\kern 1pt} {\kern 1pt} {\kern 1pt} {\kern 1pt} {\kern 1pt} {\kern 1pt} {\kern 1pt} {\kern 1pt} {\kern 1pt} {\kern 1pt} {\kern 1pt} {\kern 1pt} {\kern 1pt} {\kern 1pt} {\kern 1pt} {\rm{ = }}{\left\| {{M_{BA}} \cdot {e^{ - i \cdot t \cdot {H_G}}} \cdot {\psi _B}} \right\|^{\rm{2}}}
\end{array}
\end{equation}
To pick out the state of uncertainty, we set
\begin{equation}
{M_U} = \left( {\begin{array}{*{20}{c}}
{{M_{GU}}}&0\\
0&{{M_{BU}}}
\end{array}} \right)
\end{equation}
with
\begin{equation}
{M_{GU}} = {M_{BU}} = \left[ {\begin{array}{*{20}{c}}
0&0&0\\
0&1&0\\
0&0&0
\end{array}} \right]
\end{equation}
and then the probability of uncertain given categorizing the face as "good" equals to
\begin{equation}
\footnotesize
\begin{array}{l}
P\left( {U|G} \right){\rm{ = }}{\left\| {{M_U} \cdot {e^{ - i \cdot t \cdot H}} \cdot \psi \left( {{t_1}} \right)} \right\|^{\rm{2}}}{\rm{ = }}{\left\| {\left[ {\begin{array}{*{20}{c}}
{{M_{GU}}}&{\rm{0}}\\
{\rm{0}}&{{M_{BU}}}
\end{array}} \right] \cdot \left[ {\begin{array}{*{20}{c}}
{{e^{ - i \cdot t \cdot {H_G}}}}&{\rm{0}}\\
{\rm{0}}&{{e^{ - i \cdot t \cdot {H_B}}}}
\end{array}} \right] \cdot \left[ {\begin{array}{*{20}{c}}
{{\psi _G}}\\
0
\end{array}} \right]} \right\|^{\rm{2}}}\\
{\kern 1pt} {\kern 1pt} {\kern 1pt} {\kern 1pt} {\kern 1pt} {\kern 1pt} {\kern 1pt} {\kern 1pt} {\kern 1pt} {\kern 1pt} {\kern 1pt} {\kern 1pt} {\kern 1pt} {\kern 1pt} {\kern 1pt} {\kern 1pt} {\kern 1pt} {\kern 1pt} {\kern 1pt} {\kern 1pt} {\kern 1pt} {\kern 1pt} {\kern 1pt} {\kern 1pt} {\kern 1pt} {\kern 1pt} {\kern 1pt} {\kern 1pt} {\kern 1pt} {\kern 1pt} {\kern 1pt} {\kern 1pt} {\kern 1pt} {\kern 1pt} {\kern 1pt} {\kern 1pt} {\kern 1pt} {\kern 1pt} {\kern 1pt} {\kern 1pt} {\kern 1pt} {\kern 1pt} {\kern 1pt} {\kern 1pt} {\rm{ = }}{\left\| {{M_{GU}} \cdot {e^{ - i \cdot t \cdot {H_G}}} \cdot {\psi _G}} \right\|^{\rm{2}}}
\end{array}
\end{equation}
Then the probability of uncertainty given categorizing the face as "bad" equals to
\begin{equation}
\footnotesize
\begin{array}{l}
P\left( {U|B} \right){\rm{ = }}{\left\| {{M_U} \cdot {e^{ - i \cdot t \cdot H}} \cdot \psi \left( {{t_1}} \right)} \right\|^{\rm{2}}}{\rm{ = }}{\left\| {\left[ {\begin{array}{*{20}{c}}
{{M_{GU}}}&{\rm{0}}\\
{\rm{0}}&{{M_{BU}}}
\end{array}} \right] \cdot \left[ {\begin{array}{*{20}{c}}
{{e^{ - i \cdot t \cdot {H_G}}}}&{\rm{0}}\\
{\rm{0}}&{{e^{ - i \cdot t \cdot {H_B}}}}
\end{array}} \right] \cdot \left[ {\begin{array}{*{20}{c}}
{\rm{0}}\\
{{\psi _B}}
\end{array}} \right]} \right\|^{\rm{2}}}\\
{\kern 1pt} {\kern 1pt} {\kern 1pt} {\kern 1pt} {\kern 1pt} {\kern 1pt} {\kern 1pt} {\kern 1pt} {\kern 1pt} {\kern 1pt} {\kern 1pt} {\kern 1pt} {\kern 1pt} {\kern 1pt} {\kern 1pt} {\kern 1pt} {\kern 1pt} {\kern 1pt} {\kern 1pt} {\kern 1pt} {\kern 1pt} {\kern 1pt} {\kern 1pt} {\kern 1pt} {\kern 1pt} {\kern 1pt} {\kern 1pt} {\kern 1pt} {\kern 1pt} {\kern 1pt} {\kern 1pt} {\kern 1pt} {\kern 1pt} {\kern 1pt} {\kern 1pt} {\kern 1pt} {\kern 1pt} {\kern 1pt} {\kern 1pt} {\kern 1pt} {\kern 1pt} {\kern 1pt} {\kern 1pt} {\kern 1pt} {\rm{ = }}{\left\| {{M_{BU}} \cdot {e^{ - i \cdot t \cdot {H_B}}} \cdot {\psi _B}} \right\|^{\rm{2}}}
\end{array}
\end{equation}
Hence, the probability of finally deciding to attack equals to
\begin{equation}\label{C-D P'}
P\left( A \right) = P\left( G \right)P\left( {A|G} \right) + P\left( B \right)P\left( {A|B} \right){\rm{ + }}0.5 \left( {P\left( {U|G} \right) + P\left( {U|B} \right)} \right)
\end{equation}
Take the obtained probabilities into Eq. (\ref{C-D P'}), we can get
\begin{equation}\label{C-D p(A)}
\begin{array}{l}
P\left( A \right) = \left( {{{\left| {{\psi _{AG}}} \right|}^2} + {{\left| {{\psi _{UG}}} \right|}^2} + {{\left| {{\psi _{WG}}} \right|}^2}} \right) \cdot {\left\| {\left( {{M_{GA}} + 0.5 \cdot {M_{GU}}} \right) \cdot {e^{ - i \cdot t \cdot {H_G}}} \cdot {\psi _G}} \right\|^{\rm{2}}}\\
{\kern 1pt} {\kern 1pt} {\kern 1pt} {\kern 1pt} {\kern 1pt} {\kern 1pt} {\kern 1pt} {\kern 1pt} {\kern 1pt} {\kern 1pt} {\kern 1pt} {\kern 1pt} {\kern 1pt} {\kern 1pt} {\kern 1pt} {\kern 1pt} {\kern 1pt} {\kern 1pt} {\kern 1pt} {\kern 1pt} {\kern 1pt} {\kern 1pt} {\kern 1pt} {\kern 1pt} {\kern 1pt} {\kern 1pt} {\kern 1pt} {\kern 1pt} {\kern 1pt} {\kern 1pt} {\kern 1pt}  + \left( {{{\left| {{\psi _{AB}}} \right|}^2} + {{\left| {{\psi _{UB}}} \right|}^2} + {{\left| {{\psi _{WB}}} \right|}^2}} \right) \cdot {\left\| {\left( {{M_{BA}} + 0.5 \cdot {M_{BU}}} \right) \cdot {e^{ - i \cdot t \cdot {H_B}}} \cdot {\psi _B}} \right\|^{\rm{2}}}
\end{array}
\end{equation}

In D along condition, without a precise categorization, it is reasonable to assume that ${P\left( {U|G} \right)}$ and ${P\left( {U|B} \right)}$ will not be produced. It means that the uncertainty state will be picked out and divided in the same time rather than being produced at first and then being divided as C-D condition. To address it, we set the measure matrix as Eq. (\ref{M=MG MB}) with
\begin{equation}
{M_G} = {M_B} = \left[ {\begin{array}{*{20}{c}}
1&0&0\\
0&{{\textstyle{1 \over {\sqrt 2 }}}}&0\\
0&0&0
\end{array}} \right]
\end{equation}
To meet the sum of $P\left( A \right)$ and $P\left( G \right)$ equals to one, the measure coefficient of the uncertain state is ${{\textstyle{1 \over {\sqrt 2 }}}}$.
Then the probability of attacking in D along condition equals to
\begin{equation}\label{D-along P(A)}
\footnotesize
\begin{array}{l}
P\left( A \right){\rm{ = }}{\left\| {M \cdot {e^{ - i \cdot t \cdot H}} \cdot \psi \left( 0 \right)} \right\|^{\rm{2}}}\\
{\kern 1pt} {\kern 1pt} {\kern 1pt} {\kern 1pt} {\kern 1pt} {\kern 1pt} {\kern 1pt} {\kern 1pt} {\kern 1pt} {\kern 1pt} {\kern 1pt} {\kern 1pt} {\kern 1pt} {\kern 1pt} {\kern 1pt} {\kern 1pt} {\kern 1pt} {\kern 1pt} {\kern 1pt} {\kern 1pt} {\kern 1pt} {\kern 1pt} {\kern 1pt} {\kern 1pt} {\kern 1pt} {\kern 1pt} {\rm{ = }}{\left\| {\left[ {\begin{array}{*{20}{c}}
{{M_G}}&{\rm{0}}\\
{\rm{0}}&{{M_B}}
\end{array}} \right] \cdot \left[ {\begin{array}{*{20}{c}}
{{e^{ - i \cdot t \cdot {H_G}}}}&{\rm{0}}\\
{\rm{0}}&{{e^{ - i \cdot t \cdot {H_B}}}}
\end{array}} \right] \cdot \left[ {\begin{array}{*{20}{c}}
{\sqrt {{{\left| {{\psi _{AG}}} \right|}^2} + {{\left| {{\psi _{UG}}} \right|}^2} + {{\left| {{\psi _{WG}}} \right|}^2}}  \cdot {\psi _G}}\\
{\sqrt {{{\left| {{\psi _{AB}}} \right|}^2} + {{\left| {{\psi _{UB}}} \right|}^2} + {{\left| {{\psi _{WB}}} \right|}^2}}  \cdot {\psi _B}}
\end{array}} \right]} \right\|^{\rm{2}}}\\
{\kern 1pt} {\kern 1pt} {\kern 1pt} {\kern 1pt} {\kern 1pt} {\kern 1pt} {\kern 1pt} {\kern 1pt} {\kern 1pt} {\kern 1pt} {\kern 1pt} {\kern 1pt} {\kern 1pt} {\kern 1pt} {\kern 1pt} {\kern 1pt} {\kern 1pt} {\kern 1pt} {\kern 1pt} {\kern 1pt} {\kern 1pt} {\kern 1pt} {\kern 1pt} {\kern 1pt} {\kern 1pt} {\kern 1pt} {\rm{ = }}\left( {{{\left| {{\psi _{AG}}} \right|}^2} + {{\left| {{\psi _{UG}}} \right|}^2} + {{\left| {{\psi _{WG}}} \right|}^2}} \right) \cdot {\left\| {{M_G} \cdot {e^{ - i \cdot t \cdot {H_G}}} \cdot {\psi _G}} \right\|^{\rm{2}}}\\
{\kern 1pt} {\kern 1pt} {\kern 1pt} {\kern 1pt} {\kern 1pt} {\kern 1pt} {\kern 1pt} {\kern 1pt} {\kern 1pt} {\kern 1pt} {\kern 1pt} {\kern 1pt} {\kern 1pt} {\kern 1pt} {\kern 1pt} {\kern 1pt} {\kern 1pt} {\kern 1pt} {\kern 1pt} {\kern 1pt} {\kern 1pt} {\kern 1pt} {\kern 1pt} {\kern 1pt} {\kern 1pt} {\kern 1pt}  + \left( {{{\left| {{\psi _{AB}}} \right|}^2} + {{\left| {{\psi _{UB}}} \right|}^2} + {{\left| {{\psi _{WB}}} \right|}^2}} \right) \cdot {\left\| {{M_B} \cdot {e^{ - i \cdot t \cdot {H_G}}} \cdot {\psi _B}} \right\|^{\rm{2}}}
\end{array}
\end{equation}

Now, the probabilities of attacking in both C-D condition and D along condition have been obtained. Compare Eq. \ref{C-D p(A)} with Eq. \ref{D-along P(A)}, we can find that the two probabilities are unequal. The crucial difference of two equations is the measure matrix, which is also the key to predict the interference effect of categorization on decision making. In C-D condition, the probability of uncertain state undergoes the process of producing and dividing separately while it is produced and divided in the same time in D along condition. The inference effect of categorization acts on the uncertain state and then lead to the violation of the law of total probability.

The parameters of the model is shown in Table \ref{parameters} as follows.
\begin{table}[!h]
{\caption{the model parameters}\label{parameters}
\begin{threeparttable}
\begin{tabular*}{\columnwidth}{@{\extracolsep{\fill}}@{~~}ccccc@{~~}}
\toprule
  Parameters  & ${h_G}$ & ${h_B}$  & $P\left( A \right)$\tnote{1}  & $P\left( A \right)$\tnote{2}   \\
\midrule
  Type       & Free    & Free     &Known                &Known             \\
\bottomrule
\end{tabular*}
 \begin{tablenotes}
        \footnotesize
        \item[1] $P\left( A \right)$ equals to ${{{\left| {{\psi _{AG}}} \right|}^2} + {{\left| {{\psi _{UG}}} \right|}^2} + {{\left| {{\psi _{WG}}} \right|}^2}}$.
        \item[2] $P\left( A \right)$ equals to ${{{\left| {{\psi _{AB}}} \right|}^2} + {{\left| {{\psi _{UB}}} \right|}^2} + {{\left| {{\psi _{WB}}} \right|}^2}}$.
      \end{tablenotes}
\end{threeparttable}}
\end{table}
\subsection{Result and comparison}
Applying our QDB model to the categorization decision-making experiment, we can adjust the values of two free coefficients ${h_G}$ and ${h_B}$ and obtain a series of probabilities. Compared the obtained model results with the observed experiment results (narrow type face) in Table \ref{preworks}, the model result shown in Table \ref{result compare} is close to practical situation, which verifies the correctness and effectiveness of our model.
\begin{table}[!h]
\centering
\caption{The result of QDB model}
\label{result compare}
\begin{threeparttable}
\begin{tabular}{cccccccc}
\toprule
Literature                                                                                        &     & $P\left( G \right)$ & $P\left( {A|G} \right)$\tnote{1} & $P\left( B \right)$ & $P\left( {A|B} \right)$\tnote{2} & ${{P}_T}$     & $P\left( A \right)$   \\
\midrule
\multirow{2}{*}{\scriptsize\begin{tabular}[c]{@{}c@{}}Townsend \\ $et.al$(2000)\end{tabular}}                  & Obs & 0.17 & 0.41   & 0.83 & 0.63   & 0.59   & 0.69   \\
                                                                                                  & QDB & 0.17 & 0.41   & 0.83 & 0.6296 & 0.5923 & 0.6756 \\ \hline
\multirow{2}{*}{\scriptsize\begin{tabular}[c]{@{}c@{}}Busemeyer \\ $et.al$(2009)\end{tabular}}                 & Obs & 0.20 & 0.45   & 0.80 & 0.64   & 0.60   & 0.69   \\
                                                                                                  & QDB & 0.2  & 0.4499 & 0.80 & 0.6409 & 0.6027 & 0.6860 \\ \hline
\multirow{2}{*}{\scriptsize\begin{tabular}[c]{@{}c@{}}Wang and Busemeyer(2016)\\  Experiment 1\end{tabular}} & Obs & 0.21 & 0.41   & 0.79 & 0.58   & 0.54   & 0.59   \\
                                                                                                  & QDB & 0.21 & 0.41   & 0.79 & 0.5802 & 0.5444 & 0.6278 \\ \hline
\multirow{2}{*}{\scriptsize\begin{tabular}[c]{@{}c@{}}Wang and Busemeyer(2016) \\ Experiment 2\end{tabular}} & Obs & 0.24 & 0.37   & 0.76 & 0.61   & 0.55   & 0.60   \\
                                                                                                  & QDB & 0.24 & 0.3702 & 0.76 & 0.6104 & 0.5528 & 0.6361 \\ \hline
\multirow{2}{*}{\scriptsize\begin{tabular}[c]{@{}c@{}}Wang and Busemeyer(2016) \\ Experiment 3\end{tabular}} & Obs & 0.24 & 0.33   & 0.76 & 0.66   & 0.58   & 0.62   \\
                                                                                                  & QDB & 0.24 & 0.3296 & 0.76 & 0.6604 & 0.5810 & 0.6644 \\ \hline
\multirow{2}{*}{Average}                                                                          & Obs & 0.21 & 0.39   & 0.79 & 0.62   & 0.57   & 0.64   \\
                                                                                                  & QDB & 0.21 & 0.39   & 0.79 & 0.6205 & 0.5721 & 0.6554\\
\bottomrule
\end{tabular}
\begin{tablenotes}
        \footnotesize
        \item[1] $P\left( {A|G} \right)$ in QDB model is calculated by $P\left( {A|G} \right) = P'\left( {A|G} \right) + 0.5 \cdot P(N|G)$.
        \item[2] $P\left( {A|B} \right)$ in QDB model is calculated by $P\left( {A|B} \right) = P'\left( {A|B} \right) + 0.5 \cdot P(N|B)$.
      \end{tablenotes}
\end{threeparttable}
\end{table}

\begin{figure}[!ht]
\centering
\label{zhuzhuangtu}
\includegraphics[scale=0.53]{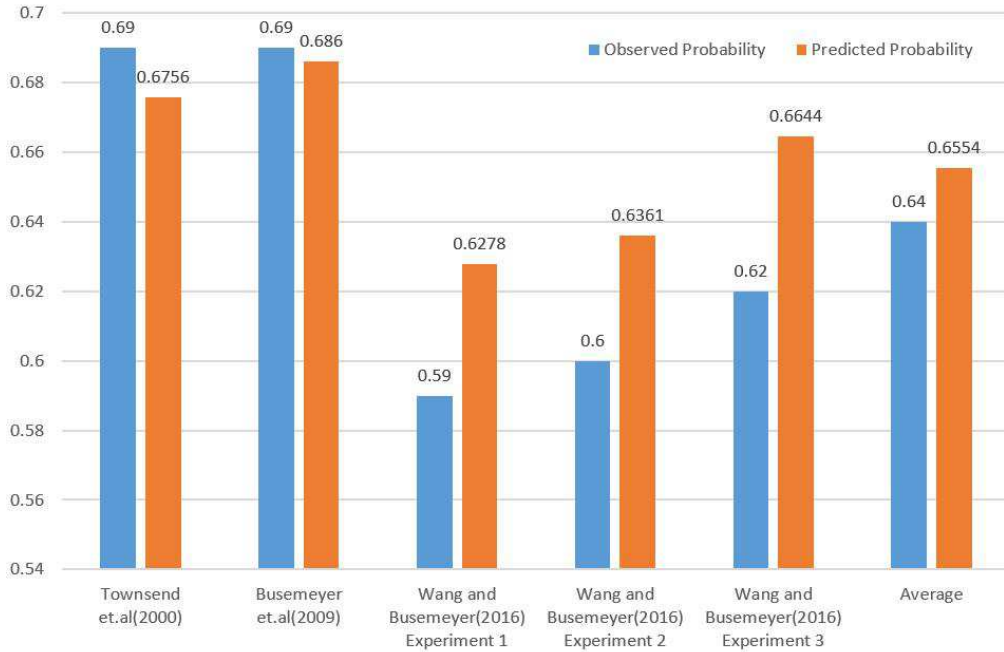}
\caption{Comparison between observed result and predicted result}\label{face}
\end{figure}

Figure \ref{zhuzhuangtu} compares the observed probability of attacking with the model predicted one. As the figure shows, the error between them is small and the average error rate is just 2.4\%. Hence, the interference effect of categorization can be well predicted.

In the following, the comparison among our QDB model, Markov BA model and quantum BAE model will be made.

Townsend et al\cite{townsend2000} proposed the Markov model to do category-decision task. The model assumes that the categorization and decision -making are two parts in the chain, namely the categorization depends only on the face while the action depends only on the categorization.
\begin{figure}[!ht]
\centering
\label{markov}
\includegraphics[scale=0.53]{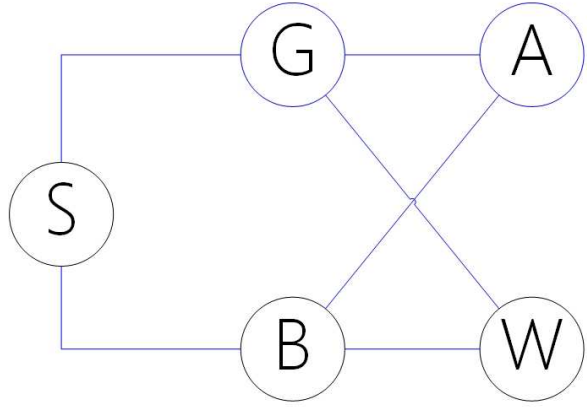}
\caption{The decision-making process of Markov BA model}\label{face}
\end{figure}
In C-D condition, the probability of attacking equals to $\phi (G) \cdot \phi (A|G)$ if categorizing the face as a good one and equals to$\phi (B) \cdot \phi (A|B)$ if categorizing the face as a bad one. In D along condition, the probability of attacking equals the probability of reaching a final state $A$ by two different paths $\phi (A) = \phi (G) \cdot \phi (A|G) + \phi (B) \cdot \phi (A|B)$. Hence, the Markov BA model follows the law of total probability, which means no interference effect will be produced.

Quantum BAE model is initially proposed by Pothos and Busemeyer(2009)\cite{Pothos2009A} and it has further developed in \cite{wang2016interference} based on quantum dynamic model. The crucial factor to produce interference effect is that a new entanglement parameter $\gamma $ is defined. As the state G and W, State B and A is assumed to be entangled in some degree, the interference effect can be produced.

The comparison result among the three models is shown in Table \ref{model compare}.
\begin{table}[!h]
\centering
\caption{Comparison among different models}
\label{model compare}
\begin{tabular}{cccc}
\toprule
Literature                                                                                        & Model       & Pt                        & P(A)                      \\
\midrule
\multirow{4}{*}{\begin{tabular}[c]{@{}c@{}}Townsend \\ $et.al$(2000)\end{tabular}}                  & Obs         & 0.59                      & 0.69                      \\
                                                                                                  & QDB         & 0.5923                    & 0.6756                    \\
                                                                                                  & Quantum BAE & 0.56                      & 0.63                      \\
                                                                                                  & Markov BA   & 0.576                     & 0.576                     \\ \hline
\multirow{4}{*}{\begin{tabular}[c]{@{}c@{}}Busemeyer \\ $et.al$(2009)\end{tabular}}                 & Obs         & 0.60                      & 0.69                      \\
                                                                                                  & QDB         & 0.6027                    & 0.6860                    \\
                                                                                                  & Quantum BAE & 0.56                      & 0.63                      \\
                                                                                                  & Markov BA   & 0.621                     & 0.621                     \\ \hline
\multirow{4}{*}{\begin{tabular}[c]{@{}c@{}}Wang and Busemeyer(2016)\\  Experiment 1\end{tabular}} & Obs         & 0.54                      & 0.59                      \\
                                                                                                  & QDB         & 0.5444                    & 0.6278                    \\
                                                                                                  & Quantum BAE & 0.5634                    & 0.6214                    \\
                                                                                                  & Markov BA   & 0.532                     & 0.532                     \\ \hline
\multirow{4}{*}{\begin{tabular}[c]{@{}c@{}}Wang and Busemeyer(2016) \\ Experiment 2\end{tabular}} & Obs         & 0.55                      & 0.60                      \\
                                                                                                  & QDB         & 0.5528                    & 0.6361                    \\
                                                                                                  & Quantum BAE & 0.6065                    & 0.6315                    \\
                                                                                                  & Markov BA   & 0.5979                    & 0.5979                    \\ \hline
\multirow{4}{*}{\begin{tabular}[c]{@{}c@{}}Wang and Busemeyer(2016) \\ Experiment 3\end{tabular}} & Obs         & 0.58                      & 0.62                      \\
                                                                                                  & QDB         & 0.5810                    & 0.6644                    \\
                                                                                                  & Quantum BAE & 0.6123                    & 0.6323                    \\
                                                                                                  & Markob BA   & 0.5316                    & 0.5316                    \\ \hline
\multirow{4}{*}{Average}                                                                          & Obs         & 0.57                      & 0.64                      \\
                                                                                                  & QDB         & 0.5721                    & 0.6554                    \\
                                                                                                  & Quantum BAE & 0.580                     & 0.629                     \\
                                                                                                  & Makov BA    & \multicolumn{1}{l}{0.572} & \multicolumn{1}{l}{0.572}\\
\bottomrule
\end{tabular}
\end{table}
Both the quantum BAE model and our QDB model can predict the interference effect while the Markov BA model could not. However, the results of QDB model are more close to the observed experiment result. In addition, our QDB model is more succinct than the quantum BAE model as it has less parameters. Based on the above, it is reasonable to conclude that the QDB model is efficient to predict the interference effect of categorization.

\section{Conclusion}\label{Conclusion}
In this paper, a quantum dynamic belief model is proposed to predict the interference effects of categorizing on decision making. Quantum probability theory is an effective tool to approach the psychology. Based on the quantum dynamic model, the conception of uncertainty is introduced to the QDB model to simulate the real process of human thinking. The model is applied to explain the categorization decision-making experiment which verifies our model's correctness and effectiveness. And the comparison with other models is also made. Admittedly, the model proposed can only predict the positive interference effect. For future study like when the interference effect will be produced still remains to be studied.

\section{Acknowledgement}
The work is partially supported by National Natural Science Foundation of China (Grant No. 61671384), Natural Science Basic Research Plan in Shaanxi Province of China (Program No. 2016JM6018), Aviation Science Foundation (Program No. 20165553036), the Fund of SAST (Program No. SAST2016083), the Seed Foundation of Innovation and Creation for Graduate Students in Northwestern Polytechnical University (Program No. Z2016122).

\bibliographystyle{model1-num-names}
\bibliography{myreference}

\begin{thebibliography}{39}
\expandafter\ifx\csname natexlab\endcsname\relax\def\natexlab#1{#1}\fi
\providecommand{\bibinfo}[2]{#2}
\ifx\xfnm\relax \def\xfnm[#1]{\unskip,\space#1}\fi
\bibitem[{Townsend et~al.(2000)Townsend, Silva, Spencer-Smith, and
  Wenger}]{townsend2000}
\bibinfo{author}{J.~T. Townsend}, \bibinfo{author}{K.~M. Silva},
  \bibinfo{author}{J.~Spencer-Smith}, \bibinfo{author}{M.~J. Wenger},
\newblock \bibinfo{title}{Exploring the relations between categorization and
  decision making with regard to realistic face stimuli},
\newblock \bibinfo{journal}{Pragmatics \& Cognition} \bibinfo{volume}{8}
  (\bibinfo{year}{2000}) \bibinfo{pages}{83--105}.
\bibitem[{Wang and Busemeyer(2016)}]{wang2016interference}
\bibinfo{author}{Z.~Wang}, \bibinfo{author}{J.~R. Busemeyer},
\newblock \bibinfo{title}{Interference effects of categorization on decision
  making},
\newblock \bibinfo{journal}{Cognition} \bibinfo{volume}{150}
  (\bibinfo{year}{2016}) \bibinfo{pages}{133--149}.
\bibitem[{Nosofsky(2015)}]{Nosofsky2015An}
\bibinfo{author}{R.~M. Nosofsky},
\newblock \bibinfo{title}{An exemplar-model account of feature inference from
  uncertain categorizations.},
\newblock \bibinfo{journal}{Journal of Experimental Psychology Learning Memory
  \& Cognition} \bibinfo{volume}{41} (\bibinfo{year}{2015}).
\bibitem[{Yu et~al.(2015)Yu, Qiu, Situ, Wang, and Long}]{yu2015enhancing}
\bibinfo{author}{F.~Yu}, \bibinfo{author}{D.~Qiu}, \bibinfo{author}{H.~Situ},
  \bibinfo{author}{X.~Wang}, \bibinfo{author}{S.~Long},
\newblock \bibinfo{title}{Enhancing user privacy in {SARG}04-based private
  database query protocols},
\newblock \bibinfo{journal}{Quantum Information Processing}
  \bibinfo{volume}{14} (\bibinfo{year}{2015}) \bibinfo{pages}{4201--4210}.
\bibitem[{Sharma et~al.(2016)Sharma, Thapliyal, Pathak, and
  Banerjee}]{Sharma2016A}
\bibinfo{author}{V.~Sharma}, \bibinfo{author}{K.~Thapliyal},
  \bibinfo{author}{A.~Pathak}, \bibinfo{author}{S.~Banerjee},
\newblock \bibinfo{title}{A comparative study of protocols for secure quantum
  communication under noisy environment: single-qubit-based protocols versus
  entangled-state-based protocols},
\newblock \bibinfo{journal}{Quantum Information Processing}
  \bibinfo{volume}{15} (\bibinfo{year}{2016}) \bibinfo{pages}{1--30}.
\bibitem[{Song et~al.(2015)Song, Qin, Wen, Wang, and Jia}]{song2015finite}
\bibinfo{author}{T.-T. Song}, \bibinfo{author}{S.-J. Qin},
  \bibinfo{author}{Q.-Y. Wen}, \bibinfo{author}{Y.-K. Wang},
  \bibinfo{author}{H.-Y. Jia},
\newblock \bibinfo{title}{Finite-key security analyses on passive decoy-state
  {QKD} protocols with different unstable sources},
\newblock \bibinfo{journal}{Scientific Reports} \bibinfo{volume}{5}
  (\bibinfo{year}{2015}) \bibinfo{pages}{15276}.
\bibitem[{Gisin and Thew(2015)}]{Gisin2015Quantum}
\bibinfo{author}{N.~Gisin}, \bibinfo{author}{R.~Thew},
\newblock \bibinfo{title}{Quantum communication},
\newblock \bibinfo{journal}{Quantum Information} \bibinfo{volume}{57}
  (\bibinfo{year}{2015}) \bibinfo{pages}{441--462}.
\bibitem[{Khrennikov(2004)}]{Khrennikov2004Information}
\bibinfo{author}{A.~Khrennikov},
\newblock \bibinfo{title}{Information dynamics in cognitive, psychological,
  social and anomalous phenomena},
\newblock \bibinfo{journal}{Fundamental Theories of Physics}
  \bibinfo{volume}{138} (\bibinfo{year}{2004}) \bibinfo{pages}{xvi,235}.
\bibitem[{Aerts and de~Bianchi(2015)}]{aerts2015unreasonable}
\bibinfo{author}{D.~Aerts}, \bibinfo{author}{M.~S. de~Bianchi},
\newblock \bibinfo{title}{The unreasonable success of quantum probability i:
  Quantum measurements as uniform fluctuations},
\newblock \bibinfo{journal}{Journal of Mathematical Psychology}
  \bibinfo{volume}{67} (\bibinfo{year}{2015}) \bibinfo{pages}{51--75}.
\bibitem[{Yukalov and Sornette(2015)}]{Yukalov2015Quantum}
\bibinfo{author}{V.~I. Yukalov}, \bibinfo{author}{D.~Sornette},
\newblock \bibinfo{title}{Quantum decision theory as quantum theory of
  measurement},
\newblock \bibinfo{journal}{Physics Letters A} \bibinfo{volume}{372}
  (\bibinfo{year}{2015}) \bibinfo{pages}{6867--6871}.
\bibitem[{Blutner et~al.(2013)Blutner, Pothos, and Bruza}]{blutner2013quantum}
\bibinfo{author}{R.~Blutner}, \bibinfo{author}{E.~M. Pothos},
  \bibinfo{author}{P.~Bruza},
\newblock \bibinfo{title}{A quantum probability perspective on borderline
  vagueness},
\newblock \bibinfo{journal}{Topics in Cognitive Science} \bibinfo{volume}{5}
  (\bibinfo{year}{2013}) \bibinfo{pages}{711--736}.
\bibitem[{Khrennikov and Haven(2009)}]{khrennikov2009quantum}
\bibinfo{author}{A.~Y. Khrennikov}, \bibinfo{author}{E.~Haven},
\newblock \bibinfo{title}{Quantum mechanics and violations of the sure-thing
  principle: the use of probability interference and other concepts},
\newblock \bibinfo{journal}{Journal of Mathematical Psychology}
  \bibinfo{volume}{53} (\bibinfo{year}{2009}) \bibinfo{pages}{378--388}.
\bibitem[{Cheon and Takahashi(2010)}]{cheon2010interference}
\bibinfo{author}{T.~Cheon}, \bibinfo{author}{T.~Takahashi},
\newblock \bibinfo{title}{Interference and inequality in quantum decision
  theory},
\newblock \bibinfo{journal}{Physics Letters A} \bibinfo{volume}{375}
  (\bibinfo{year}{2010}) \bibinfo{pages}{100--104}.
\bibitem[{Brainerd et~al.(2015)Brainerd, Wang, Reyna, and
  Nakamura}]{brainerd2015episodic}
\bibinfo{author}{C.~Brainerd}, \bibinfo{author}{Z.~Wang},
  \bibinfo{author}{V.~F. Reyna}, \bibinfo{author}{K.~Nakamura},
\newblock \bibinfo{title}{Episodic memory does not add up: Verbatim--gist
  superposition predicts violations of the additive law of probability},
\newblock \bibinfo{journal}{Journal of memory and language}
  \bibinfo{volume}{84} (\bibinfo{year}{2015}) \bibinfo{pages}{224--245}.
\bibitem[{Alnowaihi and Dhami(2016)}]{Alnowaihi2016The}
\bibinfo{author}{A.~Alnowaihi}, \bibinfo{author}{S.~Dhami},
\newblock \bibinfo{title}{The ellsberg paradox: A challenge to quantum decision
  theory?},
\newblock \bibinfo{journal}{Discussion Papers in Economics}
  (\bibinfo{year}{2016}).
\bibitem[{Busemeyer et~al.(2011)Busemeyer, Pothos, Franco, and
  Trueblood}]{busemeyer2011quantum}
\bibinfo{author}{J.~R. Busemeyer}, \bibinfo{author}{E.~M. Pothos},
  \bibinfo{author}{R.~Franco}, \bibinfo{author}{J.~S. Trueblood},
\newblock \bibinfo{title}{A quantum theoretical explanation for probability
  judgment errors.},
\newblock \bibinfo{journal}{Psychological review} \bibinfo{volume}{118}
  (\bibinfo{year}{2011}) \bibinfo{pages}{193}.
\bibitem[{Trueblood and Busemeyer(2011)}]{trueblood2011quantum}
\bibinfo{author}{J.~S. Trueblood}, \bibinfo{author}{J.~R. Busemeyer},
\newblock \bibinfo{title}{A quantum probability account of order effects in
  inference},
\newblock \bibinfo{journal}{Cognitive science} \bibinfo{volume}{35}
  (\bibinfo{year}{2011}) \bibinfo{pages}{1518--1552}.
\bibitem[{Wang and Busemeyer(2013)}]{wang2013quantum}
\bibinfo{author}{Z.~Wang}, \bibinfo{author}{J.~R. Busemeyer},
\newblock \bibinfo{title}{A quantum question order model supported by empirical
  tests of an a priori and precise prediction},
\newblock \bibinfo{journal}{Topics in Cognitive Science} \bibinfo{volume}{5}
  (\bibinfo{year}{2013}) \bibinfo{pages}{689--710}.
\bibitem[{Wang et~al.(2014)Wang, Solloway, Shiffrin, and
  Busemeyer}]{wang2014context}
\bibinfo{author}{Z.~Wang}, \bibinfo{author}{T.~Solloway},
  \bibinfo{author}{R.~M. Shiffrin}, \bibinfo{author}{J.~R. Busemeyer},
\newblock \bibinfo{title}{Context effects produced by question orders reveal
  quantum nature of human judgments},
\newblock \bibinfo{journal}{Proceedings of the National Academy of Sciences}
  \bibinfo{volume}{111} (\bibinfo{year}{2014}) \bibinfo{pages}{9431--9436}.
\bibitem[{Pothos and Busemeyer(2009)}]{Pothos2009A}
\bibinfo{author}{E.~M. Pothos}, \bibinfo{author}{J.~R. Busemeyer},
\newblock \bibinfo{title}{A quantum probability explanation for violations of
  'rational' decision theory.},
\newblock \bibinfo{journal}{Proceedings of the Royal Society B Biological
  Sciences} \bibinfo{volume}{276} (\bibinfo{year}{2009})
  \bibinfo{pages}{2171--8}.
\bibitem[{Nyman and Basieva(2011)}]{Nyman2011Quantum}
\bibinfo{author}{P.~Nyman}, \bibinfo{author}{I.~Basieva},
\newblock \bibinfo{title}{Quantum-like representation algorithm for
  trichotomous observables},
\newblock \bibinfo{journal}{International Journal of Theoretical Physics}
  \bibinfo{volume}{50} (\bibinfo{year}{2011}) \bibinfo{pages}{3864--3881}.
\bibitem[{Busemeyer and Wang(2007)}]{busemeyer2007quantum}
\bibinfo{author}{J.~R. Busemeyer}, \bibinfo{author}{Z.~Wang},
\newblock \bibinfo{title}{Quantum information processing explanation for
  interactions between inferences and decisions.},
\newblock in: \bibinfo{booktitle}{AAAI Spring Symposium: Quantum Interaction},
  pp. \bibinfo{pages}{91--97}.
\bibitem[{Busemeyer et~al.(2009)Busemeyer, Wang, and
  Lambert-Mogiliansky}]{busemeyer2009empirical}
\bibinfo{author}{J.~R. Busemeyer}, \bibinfo{author}{Z.~Wang},
  \bibinfo{author}{A.~Lambert-Mogiliansky},
\newblock \bibinfo{title}{Empirical comparison of markov and quantum models of
  decision making},
\newblock \bibinfo{journal}{Journal of Mathematical Psychology}
  \bibinfo{volume}{53} (\bibinfo{year}{2009}) \bibinfo{pages}{423--433}.
\bibitem[{Nyman(2011)}]{Nyman2011On}
\bibinfo{author}{P.~Nyman},
\newblock \bibinfo{title}{On the consistency of the quantum-like representation
  algorithm for hyperbolic interference},
\newblock \bibinfo{journal}{Advances in Applied Clifford Algebras}
  \bibinfo{volume}{21} (\bibinfo{year}{2011}) \bibinfo{pages}{799--811}.
\bibitem[{Yukalov and Sornette(2011)}]{Yukalov2011Decision}
\bibinfo{author}{V.~I. Yukalov}, \bibinfo{author}{D.~Sornette},
\newblock \bibinfo{title}{Decision theory with prospect interference and
  entanglement},
\newblock \bibinfo{journal}{Theory and Decision} \bibinfo{volume}{70}
  (\bibinfo{year}{2011}) \bibinfo{pages}{283--328}.
\bibitem[{Moreira and Wichert(2016)}]{Moreira2016Quantum}
\bibinfo{author}{C.~Moreira}, \bibinfo{author}{A.~Wichert},
\newblock \bibinfo{title}{Quantum-like bayesian networks for modeling decision
  making.},
\newblock \bibinfo{journal}{Frontiers in Psychology} \bibinfo{volume}{7}
  (\bibinfo{year}{2016}).
\bibitem[{Busemeyer and Pothos(2012)}]{busemeyer2012social}
\bibinfo{author}{J.~R. Busemeyer}, \bibinfo{author}{E.~M. Pothos},
\newblock \bibinfo{title}{Social projection and a quantum approach for behavior
  in prisoner's dilemma},
\newblock \bibinfo{journal}{Psychological Inquiry} \bibinfo{volume}{23}
  (\bibinfo{year}{2012}) \bibinfo{pages}{28--34}.
\bibitem[{Chen et~al.(2003)Chen, Ang, Kiang, Kwek, and Lo}]{chen2003quantum}
\bibinfo{author}{L.~Chen}, \bibinfo{author}{H.~Ang},
  \bibinfo{author}{D.~Kiang}, \bibinfo{author}{L.~Kwek},
  \bibinfo{author}{C.~Lo},
\newblock \bibinfo{title}{Quantum prisoner dilemma under decoherence},
\newblock \bibinfo{journal}{Physics Letters A} \bibinfo{volume}{316}
  (\bibinfo{year}{2003}) \bibinfo{pages}{317--323}.
\bibitem[{Makowski(2009)}]{Makowski2009Transitivity}
\bibinfo{author}{M.~Makowski},
\newblock \bibinfo{title}{Transitivity vs. intransitivity in decision making
  process ¨c an example in quantum game theory},
\newblock \bibinfo{journal}{Physics Letters A} \bibinfo{volume}{373}
  (\bibinfo{year}{2009}) \bibinfo{pages}{2125--2130}.
\bibitem[{Alonso-Sanz et~al.(2015)Alonso-Sanz, Carvalho, and
  Situ}]{Alonso2015A}
\bibinfo{author}{R.~Alonso-Sanz}, \bibinfo{author}{M.~Carvalho},
  \bibinfo{author}{H.~Situ},
\newblock \bibinfo{title}{A quantum relativistic prisoner¡¯s dilemma cellular
  automaton},
\newblock \bibinfo{journal}{International Journal of Theoretical Physics}
  \bibinfo{volume}{470} (\bibinfo{year}{2015}) \bibinfo{pages}{1--14}.
\bibitem[{Alonso-Sanz and Situ(2016)}]{Alonso2016On}
\bibinfo{author}{R.~Alonso-Sanz}, \bibinfo{author}{H.~Situ},
\newblock \bibinfo{title}{On the effect of quantum noise in a
  quantum-relativistic prisoner¡¯s dilemma cellular automaton},
\newblock \bibinfo{journal}{International Journal of Theoretical Physics}
  \bibinfo{volume}{55} (\bibinfo{year}{2016}) \bibinfo{pages}{5265--5279}.
\bibitem[{Busemeyer et~al.(2006)Busemeyer, Wang, and
  Townsend}]{busemeyer2006quantum}
\bibinfo{author}{J.~R. Busemeyer}, \bibinfo{author}{Z.~Wang},
  \bibinfo{author}{J.~T. Townsend},
\newblock \bibinfo{title}{Quantum dynamics of human decision-making},
\newblock \bibinfo{journal}{Journal of Mathematical Psychology}
  \bibinfo{volume}{50} (\bibinfo{year}{2006}) \bibinfo{pages}{220--241}.
\bibitem[{Deng(2015)}]{deng2015Generalized}
\bibinfo{author}{Y.~Deng},
\newblock \bibinfo{title}{Generalized evidence theory},
\newblock \bibinfo{journal}{Applied Intelligence} \bibinfo{volume}{43}
  (\bibinfo{year}{2015}) \bibinfo{pages}{530--543}.
\bibitem[{Du et~al.(2015)Du, Gao, Liu, Zheng, and Wang}]{Du2015832}
\bibinfo{author}{W.~Du}, \bibinfo{author}{Y.~Gao}, \bibinfo{author}{C.~Liu},
  \bibinfo{author}{Z.~Zheng}, \bibinfo{author}{Z.~Wang},
\newblock \bibinfo{title}{Adequate is better: particle swarm optimization with
  limited-information},
\newblock \bibinfo{journal}{Applied Mathematics and Computation}
  \bibinfo{volume}{268} (\bibinfo{year}{2015}) \bibinfo{pages}{832 -- 838}.
\bibitem[{Akyar(2016)}]{2016AkyarFuzzy}
\bibinfo{author}{H.~Akyar},
\newblock \bibinfo{title}{{Fuzzy Risk Analysis for a Production System Based on
  the Nagel Point of a Triangle}},
\newblock \bibinfo{journal}{{MATHEMATICAL PROBLEMS IN ENGINEERING}}
  \bibinfo{volume}{{3080679}} (\bibinfo{year}{{2016}})
  \bibinfo{pages}{{DOI:10.1155/2016/3080679}}.
\bibitem[{He et~al.(2012)He, Hu, Guan, Han, and Deng}]{You2012New}
\bibinfo{author}{Y.~He}, \bibinfo{author}{L.~Hu}, \bibinfo{author}{X.~Guan},
  \bibinfo{author}{D.~Han}, \bibinfo{author}{Y.~Deng},
\newblock \bibinfo{title}{New conflict representation model in generalized
  power space},
\newblock \bibinfo{journal}{Journal of Systems Engineering and Electronics}
  \bibinfo{volume}{23} (\bibinfo{year}{2012}) \bibinfo{pages}{1--9}.
\bibitem[{Deng(2016)}]{dengentropy}
\bibinfo{author}{Y.~Deng},
\newblock \bibinfo{title}{Deng entropy},
\newblock \bibinfo{journal}{Chaos, Solitons \& Fractals} \bibinfo{volume}{91}
  (\bibinfo{year}{2016}) \bibinfo{pages}{549--553}.
\bibitem[{Vourdas(2014)}]{Vourdas2014Quantum}
\bibinfo{author}{A.~Vourdas},
\newblock \bibinfo{title}{Quantum probabilities as dempster-shafer
  probabilities in the lattice of subspaces},
\newblock \bibinfo{journal}{Journal of Mathematical Physics}
  \bibinfo{volume}{55} (\bibinfo{year}{2014}) \bibinfo{pages}{823}.
\bibitem[{Smets and Kennes(1994)}]{Smets1994The}
\bibinfo{author}{P.~Smets}, \bibinfo{author}{R.~Kennes},
\newblock \bibinfo{title}{The transferable belief model},
\newblock \bibinfo{journal}{Artificial Intelligence} \bibinfo{volume}{66}
  (\bibinfo{year}{1994}) \bibinfo{pages}{191--234}.

\end{thebibliography}

\end{document}